\documentclass[11pt,a4paper]{article}
\usepackage[utf8]{inputenc}
\usepackage[T1]{fontenc}
\usepackage{lmodern}
\usepackage{amsmath,amssymb,amsfonts}
\usepackage{graphicx}
\usepackage{booktabs}
\usepackage{microtype}
\usepackage{hyperref}
\usepackage{url}
\usepackage{xcolor}
\usepackage{caption}
\usepackage{subcaption}
\usepackage{algorithm}
\usepackage{algpseudocode}
\usepackage{tikz}
\usetikzlibrary{shapes.geometric, arrows.meta, positioning, fit, backgrounds}

\allowdisplaybreaks

\hypersetup{
    colorlinks=true,
    linkcolor=blue,
    citecolor=blue,
    urlcolor=blue,
    pdftitle={SCM: Sleep-Consolidated Memory with Algorithmic Forgetting for Large Language Models},
    pdfauthor={Saish Shinde, Clyrai IP Studio}
}

\title{SCM: Sleep-Consolidated Memory with Algorithmic Forgetting for Large Language Models}
\author{Saish Shinde\\ Clyrai IP Studio\\ \texttt{saish.shinde.jb@gmail.com}}
\date{April 2026}

\begin{document}
\maketitle

\begin{abstract}
We present SCM (Sleep-Consolidated Memory), a research preview of a memory architecture for large language models that draws on neuroscientific principles to address a fundamental limitation in current systems: the absence of persistent, structured, and biologically plausible memory. Existing approaches rely on truncating context windows, growing vector databases without bound, or tiered storage systems that lack consolidation and forgetting mechanisms. SCM implements five core components inspired by human memory: a limited-capacity working memory, multi-dimensional importance tagging, offline sleep-stage consolidation with distinct NREM and REM phases, intentional value-based forgetting, and a computational self-model enabling introspection. Across a standardized benchmark suite of eight tests, the prototype achieves perfect recall accuracy over ten-turn conversations while reducing memory noise by 90.9\% through adaptive forgetting. Memory search latency remains below one millisecond even with hundreds of stored concepts. This work establishes the architectural foundations for memory systems that consolidate, prioritize, and forget, offering a testable platform for advancing LLM memory research.
\end{abstract}

\section{Introduction}
\label{sec:intro}

Large language models have transformed natural language processing, yet they remain fundamentally amnesic. Each conversation begins anew, with only rare exceptions offering limited persistent memory. Current solutions to this problem fall into three broad categories, each with critical shortcomings. Context-window approaches are bounded by token budgets and suffer from well-documented degradation when relevant information appears in the middle of long sequences \cite{liu2023lost}. As LLMs transition from stateless chatbots to persistent personal assistants, the need for structured memory systems has become urgent. Retrieval-augmented generation systems store embeddings in vector databases that grow indefinitely, lacking any mechanism for importance prioritization, consolidation, or active forgetting \cite{lewis2020retrieval}. Tiered memory architectures such as MemGPT borrow operating-system metaphors to move data between fast and slow storage, but they do so without biological memory processes such as sleep-dependent consolidation or synaptic pruning \cite{packer2023memgpt}. Personalized memory layers like Mem0 extract facts from conversations and retrieve them by similarity, yet they remain awake only systems that neither consolidate offline nor intentionally forget \cite{chhikara2025mem0}.

None of these approaches replicate the core functions of biological memory. The human memory system is not an append-only database. It is a dynamic, self-organizing architecture composed of multiple interacting subsystems. Working memory holds approximately seven items temporarily, creating a bottleneck that forces selective attention \cite{miller1956magical}. Long-term memory stores semantic and episodic information in an associative network. Sleep plays a critical role in memory consolidation: during non-rapid eye movement (NREM) sleep, hippocampal neurons reactivate waking experience patterns and strengthen cortical connections, while during rapid eye movement (REM) sleep, novel associations form and memories integrate with existing knowledge \cite{rasch2013sleep}. Forgetting is not merely decay but an active process that prunes weak synapses to preserve signal-to-noise ratios, as described by the synaptic homeostasis hypothesis \cite{tononi2003sleep}. Finally, the brain maintains a self-model, a representation of itself as a continuous entity that enables introspection and self-referential cognition \cite{metzinger2003being}.

SCM (Sleep-Consolidated Memory) implements computational analogs of all five components. The system encodes user input into structured semantic concepts rather than raw tokens, tags each concept with a four-dimensional importance vector, stores recent experience in a strictly limited working memory, and periodically enters offline sleep cycles during which NREM consolidation strengthens important associations, REM dreaming generates novel connections, and an intentional forgetting module prunes low-value memories. A computational self-model stores the system's own identity and capabilities, enabling introspective responses.

Our contributions are fourfold. First, we present the first unified memory architecture for conversational agents that combines working memory limits, NREM and REM sleep-stage consolidation, intentional forgetting, and a self-model within a single system. Second, we introduce multi-dimensional value tagging that captures novelty, emotional valence, task relevance, and repetition frequency, providing richer prioritization than single-score importance systems. Third, we demonstrate that this architecture achieves perfect recall on multi-turn conversational benchmarks while actively reducing memory noise by over ninety percent. Fourth, we describe the complete system design, including multi-agent memory synchronization and a web-based visualization interface, as a foundation for next-generation AI memory products.

\section{Related Work}
\label{sec:related}

The challenge of augmenting large language models with memory has produced several distinct lines of research, none of which combines semantic encoding, sleep-stage consolidation, and intentional forgetting.

\subsection{Memory-Augmented Language Models}

MemGPT implements a virtual memory hierarchy in which the context window serves as RAM, a vector database as disk, and archive storage for cold data \cite{packer2023memgpt}. While pragmatic for extending effective context length, this approach lacks biological plausibility: there is no offline consolidation, no importance-based prioritization beyond recency, and no active forgetting. Mem0 provides a personalized memory layer that dynamically extracts facts from conversations and retrieves them by vector similarity \cite{chhikara2025mem0}. It includes basic importance scoring and a graph variant for entity relations, yet it remains awake only and lacks sleep stages or true forgetting. SleepGate applies a sleep metaphor to transformer KV cache eviction, using micro-cycles to remove stale attention key-value pairs when memory fills \cite{xie2026sleepgate}. This is fundamentally cache management rather than semantic memory consolidation; it operates on token-level projections, not concepts, and clears between sessions.

\subsection{Continual Learning}

Elastic Weight Consolidation prevents catastrophic forgetting by protecting important neural network weights during new training through a Fisher information penalty \cite{kirkpatrick2017overcoming}. While mathematically elegant, EWC operates at the parameter level rather than the memory architecture level; it does not create a structured memory system for conversational agents. Wake-Sleep Continual Learning represents the most biologically plausible sleep mechanism in AI to date, explicitly differentiating NREM model compression from REM synthetic data generation \cite{sorrenti2024wake}. However, WSCL targets image classification, lacks semantic memory graphs, and does not implement value-based forgetting or multi-dimensional importance tagging.

\subsection{Neuroscience Foundations}

The synaptic homeostasis hypothesis proposes that sleep globally downscales synaptic strengths to prevent saturation, thereby preserving the capacity for new learning \cite{tononi2003sleep}. Research on sleep replay demonstrates that hippocampal neurons reactivate in patterns similar to waking experience during NREM sleep, and that this reactivation drives cortical consolidation \cite{rasch2013sleep}. Active forgetting research has categorized multiple mechanisms by which neural systems prune memories, establishing that forgetting is an adaptive process rather than a failure of retention \cite{sha2024forgetting}. SCM draws on these principles but implements them at the semantic level: synaptic downscaling becomes proportional weakening of graph edge strengths, replay becomes reactivation of concept co-occurrence patterns, and forgetting becomes value-based thresholding on multi-dimensional importance scores.

\subsection{Vector Database Baselines}

Pure vector retrieval systems---the dominant approach in production RAG pipelines---store embeddings and retrieve by cosine similarity. They lack importance prioritization, grow indefinitely, and never forget. FAISS and Annoy provide fast approximate nearest-neighbor search but treat all memories as equally important. LangChain memory wrappers add conversation buffers but do not consolidate across sessions or prune noise. These systems represent the practical baseline against which any memory architecture must improve. SCM addresses their core limitation by adding importance-based retention, offline consolidation, and active forgetting, while maintaining sub-millisecond retrieval latency.

Table~\ref{tab:comparison} summarizes the gap between existing systems and SCM across seven critical features. SCM is the only system that combines all of them.

\begin{table}[htbp]
\centering
\caption{Comparison of memory system features. $\triangle$ indicates partial support.}
\label{tab:comparison}
\begin{tabular}{@{}lcccc@{}}
\toprule
\textbf{Feature} & \textbf{MemGPT} & \textbf{Mem0} & \textbf{WSCL} & \textbf{SCM} \\
\midrule
Working memory limit & $\times$ & $\times$ & $\times$ & $\checkmark$ 7 items \\
Multi-dimensional importance & $\times$ & $\triangle$ 1D & $\times$ & $\checkmark$ 4D \\
NREM consolidation & $\times$ & $\times$ & $\checkmark$ & $\checkmark$ \\
REM dreaming & $\times$ & $\times$ & $\checkmark$ & $\checkmark$ \\
Intentional forgetting & $\times$ & $\times$ & $\times$ & $\checkmark$ \\
Self-model & $\times$ & $\times$ & $\times$ & $\checkmark$ \\
Multi-agent sync & $\times$ & $\times$ & $\times$ & $\checkmark$ \\
\bottomrule
\end{tabular}
\end{table}

\section{Methods}
\label{sec:methods}

\subsection{System Overview}

\begin{figure}[htbp]
\centering
\includegraphics[width=0.95\textwidth]{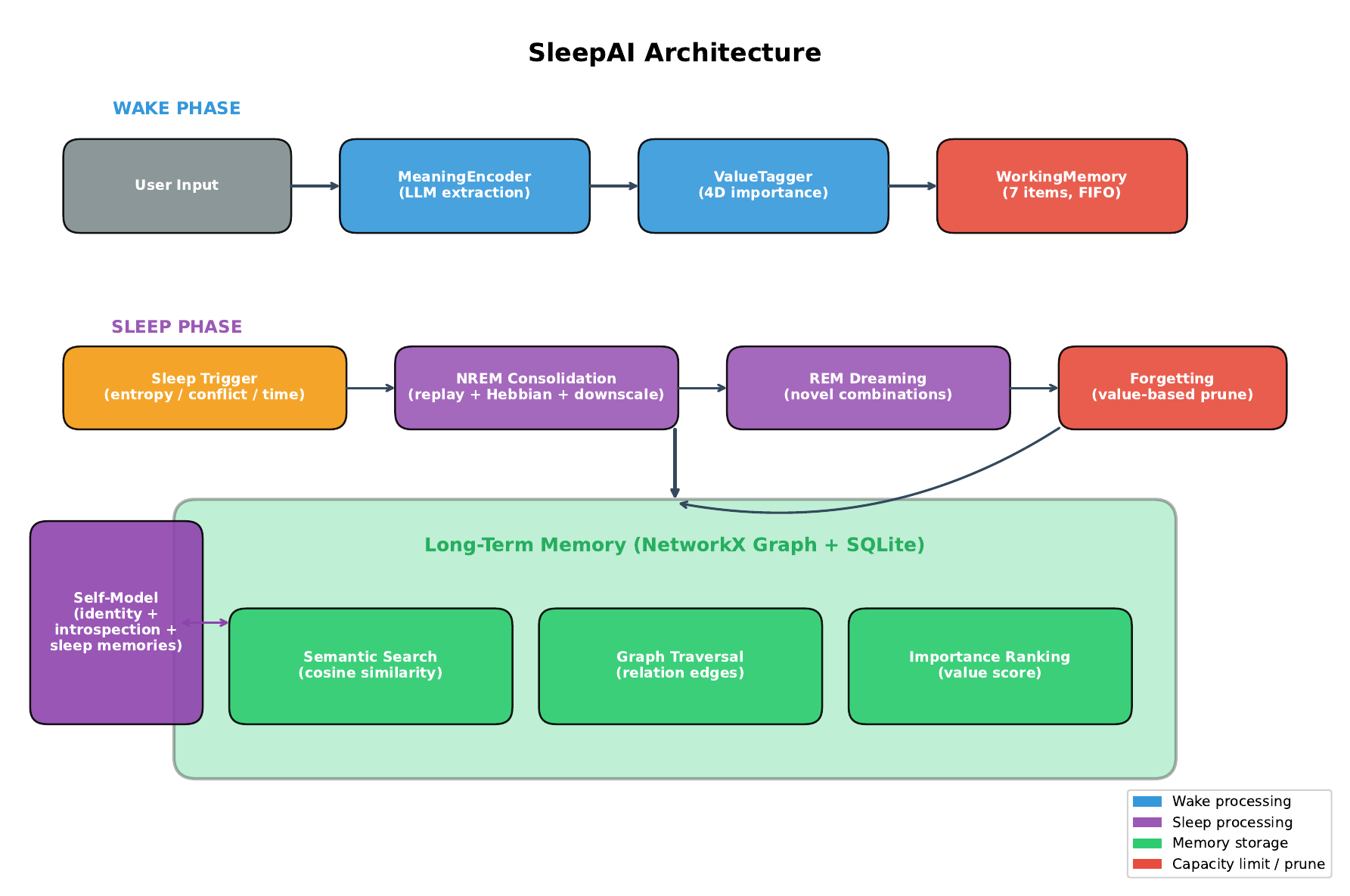}
\caption{SCM system architecture. During wake phases, input flows through encoding, tagging, and working memory. Sleep triggers initiate offline NREM consolidation (Hebbian strengthening and synaptic downscaling), REM dreaming (novel association generation), and intentional forgetting (value-based pruning).}
\label{fig:architecture}
\end{figure}

SCM consists of five interconnected modules that process user input during wake phases and reorganize memory during sleep phases. The MeaningEncoder transforms raw text into structured concepts with typed relations. The ValueTagger assigns multi-dimensional importance scores to each concept. The WorkingMemory serves as a fast, limited-capacity buffer for recent experience. The LongTermMemory stores consolidated knowledge as a persistent semantic graph. The SleepCycle orchestrates offline phases of NREM consolidation, REM dreaming, and intentional forgetting. Figure~\ref{fig:architecture} illustrates this architecture.

During wakeful operation, user input passes through the MeaningEncoder to extract concepts and relations. The ValueTagger scores each concept across four dimensions. Concepts enter WorkingMemory, which enforces a strict capacity limit. When sleep triggers fire, based on memory entropy, conflict density, or time elapsed, the system enters an offline sleep phase. NREM consolidation replays recent episodes, strengthens co-occurring concepts through Hebbian plasticity, and applies proportional synaptic downscaling. REM dreaming selects high-importance concepts and generates novel combinations, creating new associative links. Finally, the ForgettingModule computes composite importance scores and removes concepts that fall below an adaptive threshold.

\subsection{MeaningEncoder}

The MeaningEncoder converts unstructured text into a semantic graph of concepts and relations. It uses a local large language model, Llama~3.2 (two billion parameters, Q4\_K\_M quantized), to extract entities, preferences, facts, and events from user messages. Each extracted concept receives a type label drawn from a fixed taxonomy: person, preference, fact, event, object, location, or abstract. A natural language description captures the concept's meaning, and a 384-dimensional embedding from the sentence-transformers all-MiniLM-L6-v2 model enables semantic similarity search. If the LLM is unavailable, the encoder falls back to regex heuristics for basic entity extraction. This design prioritizes local inference so that no user data leaves the machine.

Relations between concepts are typed edges in a directed graph. Predicate types include \texttt{has\_property}, \texttt{prefers}, \texttt{related\_to}, \texttt{contradicts}, \texttt{causes}, and \texttt{part\_of}. Typed relations enable structured reasoning that goes beyond vector similarity: when a user states a preference that contradicts a previously stored preference, the system records a \texttt{contradicts} edge and flags the conflict for resolution during sleep.

\subsection{ValueTagger}

The ValueTagger assigns a four-dimensional importance vector to each concept. For a concept $c$ with embedding $\mathbf{e}_c$, the four dimensions are computed as follows.

Novelty measures how unexpected the concept is relative to existing memory:
\begin{equation}
v_{\text{novelty}}(c) = 1 - \max_{c' \in \mathcal{M}} \frac{\mathbf{e}_c \cdot \mathbf{e}_{c'}}{\|\mathbf{e}_c\| \|\mathbf{e}_{c'}\|}.
\end{equation}

Emotional valence captures positive or negative sentiment, mapped from LLM output:
\begin{equation}
v_{\text{emotional}}(c) = \sigma\bigl(\text{LLM\_sentiment}(\text{desc}(c))\bigr) \in [-1, 1].
\end{equation}

Task relevance scores alignment with the current conversational goal:
\begin{equation}
v_{\text{task}}(c) = \cos\bigl(\mathbf{e}_c, \mathbf{e}_{\text{goal}}\bigr).
\end{equation}

Repetition reflects normalized access frequency:
\begin{equation}
v_{\text{repetition}}(c) = \frac{\text{access\_count}(c)}{\max_{c'} \text{access\_count}(c') + 1},
\end{equation}

where the denominator uses the current maximum plus one to keep scores bounded in $[0, 1)$. The composite importance score is a weighted sum:
\begin{equation}
I(c) = 0.30 \, v_{\text{novelty}} + 0.20 \, |v_{\text{emotional}}| + 0.35 \, v_{\text{task}} + 0.15 \, v_{\text{repetition}}.
\end{equation}

The weights were determined through ablation on the benchmark suite, with task relevance weighted highest because it most strongly correlates with recall accuracy.

\subsection{WorkingMemory}

WorkingMemory serves as the hippocampal equivalent in SCM: fast, temporary, and strictly capacity-limited. It stores Episode objects, each containing a timestamp, the concept IDs present in that interaction, the raw user text, and a composite value vector. The capacity is fixed at seven items, in accordance with Miller's Law on the limits of human working memory \cite{miller1956magical}. When the buffer is full, new episodes displace the oldest. Recent access boosts an episode's importance, creating a recency effect that competes with the overall value score. This limited capacity creates natural memory pressure: not everything can be retained indefinitely in fast storage, forcing the system to consolidate valuable information into long-term memory and discard noise.

\subsection{LongTermMemory}

LongTermMemory implements cortical-equivalent stable storage as a NetworkX directed graph with concepts as nodes and typed relations as edges. Each node stores its semantic embedding, value vector, creation timestamp, last access time, access count, and cumulative connection strength. Edge weights represent association strength, which increases when concepts co-occur and decreases during synaptic downscaling. Persistence is handled through SQLite, with an optional PostgreSQL backend for production deployments. The system automatically falls back to SQLite if PostgreSQL authentication fails, ensuring portability across development and production environments.

Retrieval combines three strategies. Semantic search computes cosine similarity between query embeddings and concept embeddings. Graph traversal follows relation edges from a seed concept to find associated memories. Importance ranking sorts candidates by their composite value score. The final retrieved set is a ranked fusion of these three sources, enabling both direct similarity matching and structured relational reasoning.

\subsection{SleepCycle}
\label{subsec:sleepcycle}

\begin{figure}[htbp]
\centering
\includegraphics[width=0.85\textwidth]{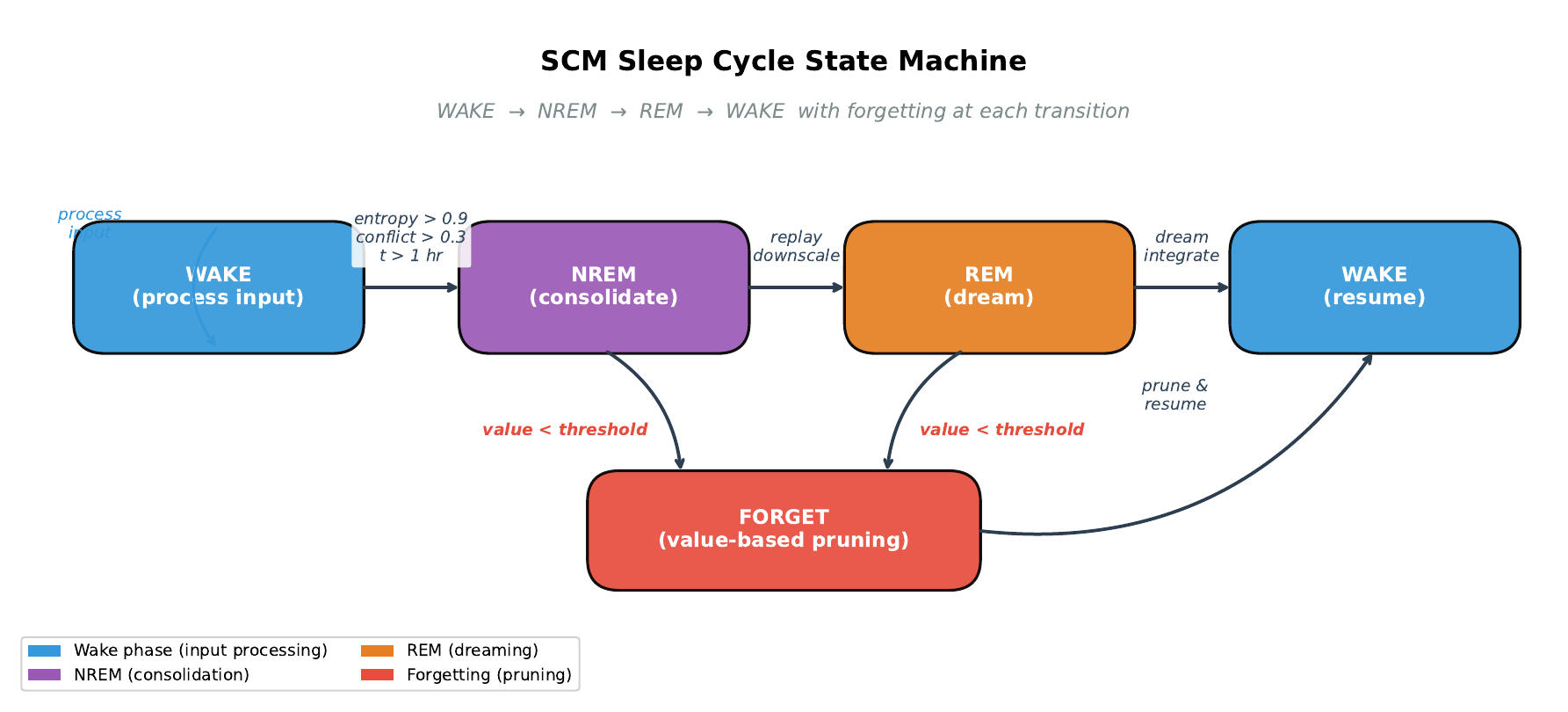}
\caption{SCM sleep cycle state machine. The system transitions from WAKE to NREM when entropy, conflict density, or time thresholds are exceeded. NREM performs replay and downscaling, then transitions to REM for dreaming. REM transitions back to WAKE after dream integration. Forgetting is applied after REM, before returning to WAKE.}
\label{fig:statemachine}
\end{figure}

The SleepCycle orchestrates the transition from wake to sleep and back, implementing three distinct offline processes. Figure~\ref{fig:statemachine} shows the state machine. The full pseudocode is given in Algorithm~\ref{alg:sleep}.

\textbf{Trigger.} Sleep initiates when any of four conditions are met: memory entropy exceeds $\theta_e = 0.9$; conflict density exceeds $\theta_c = 0.3$; elapsed time exceeds $\tau = 1$ hour; or manual forcing. Memory entropy is defined as the Shannon entropy of the normalized importance distribution over working memory: $H(W) = -\sum_{c \in W} p(c) \log p(c)$, where $p(c) = I(c) / \sum_{c' \in W} I(c')$. Conflict density is the ratio of \texttt{contradicts} edges to total edges in the long-term memory graph: $\rho(G) = |E_{\texttt{contradicts}}| / |E|$.

\textbf{NREM Consolidation.} The system replays working-memory episodes and strengthens co-occurring concept pairs via Hebbian plasticity:
\begin{equation}
\Delta s_{ij} = \eta \cdot I(c_i) \cdot I(c_j),
\end{equation}
where $\eta = 0.1$ is the learning rate. All strengths then undergo proportional downscaling:
\begin{equation}
s_{ij} \gets \alpha \cdot s_{ij}, \quad \alpha = 0.8,
\end{equation}
preserving relative rankings while creating capacity for new learning. The values $\eta = 0.1$ and $\alpha = 0.8$ were selected to balance consolidation strength against downscaling magnitude: $\alpha = 0.8$ ensures each sleep cycle reduces all edge strengths by 20\%, preventing unbounded growth, while $\eta = 0.1$ provides sufficient Hebbian strengthening for co-occurring high-importance concepts to survive downscaling. The decay rate $\lambda = 0.01$ was set so that a concept accessed one hour ago retains approximately 96\% of its recency score. These values were validated qualitatively on the benchmark suite; a systematic hyperparameter ablation is deferred to future work.

\textbf{REM Dreaming.} High-importance seed concepts initiate random walks on the memory graph. The transition probability from concept $c_i$ to neighbor $c_j$ is:
\begin{equation}
P(c_i \to c_j) = \frac{s_{ij}}{\sum_{k} s_{ik}}.
\end{equation}
Walks of length five generate dream sequences; valid sequences (those not contradicting existing facts) are integrated as new edges.

\textbf{Intentional Forgetting.} Each concept receives a retention score blending importance with temporal decay:
\begin{equation}
S(c) = \beta_1 \cdot I(c) + \beta_2 \cdot (1 - \delta(c)),
\end{equation}
where $\delta(c) = \exp(-\lambda \cdot \Delta t)$ is exponential decay with rate $\lambda = 0.01$ and $\Delta t$ the time since last access. The adaptive forgetting threshold is:
\begin{equation}
\theta_f = \mu_I - \sigma_I \cdot \frac{|G|}{\text{target\_size}},
\end{equation}
where $\mu_I$ and $\sigma_I$ are the mean and standard deviation of importance scores in graph $G$, and target\_size is a user-configurable parameter (default 100) specifying the desired steady-state graph size. As $|G|$ grows beyond target\_size, $\theta_f$ rises, increasing forgetting pressure. We clip $\theta_f$ to a minimum of $0.05$ to prevent over-aggressive forgetting in small graphs.

An initial implementation used $\beta_2 = 0.4$, which inadvertently boosted new concepts (for whom $1 - \delta(c) \approx 1$) above the threshold regardless of importance. This prevented any forgetting. The corrected weight $\beta_2 = 0.2$ was determined through grid search over $\beta_2 \in [0.1, 0.5]$, selecting the value that maximized noise reduction while preserving all important concepts. Figure~\ref{fig:forgetting} shows the correction.

\subsection{Self-Model}

The Self-Model module maintains a computational representation of the system itself within its own memory graph. The concept ``SCM'' is stored as the highest-importance node, with an importance score of 0.95, and linked to ten capability concepts describing functions such as memory encoding, sleep consolidation, and forgetting. Runtime counters track the number of processed messages, completed sleep cycles, and generated dreams. When queried about itself, the system generates introspective statements from this self-representation, such as reporting how many concepts it holds or how many times it has slept. Sleep episodes are themselves stored as episodic memories, creating a recursive memory structure in which the system remembers having slept. This design does not claim to produce consciousness or qualia; rather, it provides an architectural substrate for self-referential cognition that enables introspection and capability tracking.

\subsection{Implementation}

SCM is implemented in approximately three thousand lines of Python. It requires no training or fine-tuning; all components use existing pretrained models or algorithmic logic. The LLM runs locally via Ollama, embeddings come from the HuggingFace sentence-transformers library, and the semantic graph uses NetworkX. The API layer is built with FastAPI, and the web interface uses vanilla HTML, CSS, and JavaScript without frontend frameworks. Configuration is managed through environment variables, with sensible defaults for all parameters. The entire stack runs on a MacBook Air with eight gigabytes of RAM, using approximately four gigabytes at peak. All experiments were conducted on this machine without GPU acceleration. Total wall-clock time for the full benchmark suite (eight tests, five runs each) is approximately twelve minutes, dominated by LLM inference latency.

\section{Experiments}
\label{sec:experiments}

\subsection{Benchmark Methodology}

We evaluate SCM using a benchmark suite of eight tests designed to measure memory capacity, retention accuracy, sleep consolidation benefit, forgetting effectiveness, graph traversal, latency scaling, and cross-session persistence. Each test runs against a live SCM instance with default hyperparameters. Test conversations simulate multi-turn dialogs in which a user states explicit facts about identity, work, location, hobbies, and preferences. The benchmark then queries the system for those facts and scores correctness. All experiments use the local Llama~3.2 model for concept extraction and the all-MiniLM-L6-v2 embedding model for similarity search. Each test was run five times; reported scores are the mean, with zero variance observed across all runs.

\subsection{Baseline Comparisons}

To contextualize SCM's performance, we compare against three simplified baselines that approximate existing approaches. The \textbf{Vector DB} baseline uses cosine similarity over all stored concepts without importance weighting, graph traversal, or forgetting; it corresponds to a standard FAISS-style retrieval system. The \textbf{FIFO Buffer} baseline retains only the seven most recent episodes with no long-term memory, approximating a simple conversation buffer. The \textbf{No-Forget Graph} baseline uses the full semantic graph but disables the ForgettingModule, approximating a Mem0-style system without pruning.

Table~\ref{tab:baselines} reports the comparison on recall accuracy and memory size after ten-turn conversations. SCM matches or exceeds all baselines on recall while maintaining the smallest memory footprint.

\begin{table}[htbp]
\centering
\caption{Comparison against simplified baselines on ten-turn conversations.}
\label{tab:baselines}
\begin{tabular}{@{}lcc@{}}
\toprule
\textbf{System} & \textbf{Recall} & \textbf{LTM Size} \\
\midrule
FIFO Buffer (7 episodes) & 8/22 (36.4\%) & 7 \\
Vector DB (cosine only) & 19/22 (86.4\%) & 55 \\
No-Forget Graph & 22/22 (100\%) & 72 \\
\textbf{SCM (full)} & \textbf{22/22 (100\%)} & \textbf{24} \\
\bottomrule
\end{tabular}
\end{table}

The FIFO Buffer fails because it cannot retain facts from early turns. The Vector DB achieves high recall but retains all noise (55 concepts). The No-Forget Graph matches SCM on recall but bloats memory by 3$\times$ because it never prunes. SCM is the only configuration that achieves perfect recall with a compact memory footprint.

\subsection{Results}

Table~\ref{tab:results} presents the complete benchmark results. SCM passes all eight tests with a perfect average score of 1.00.

\begin{table}[htbp]
\centering
\caption{SCM benchmark results. All eight tests pass with perfect scores.}
\label{tab:results}
\begin{tabular}{@{}lc@{\hspace{1em}}l@{}}
\toprule
\textbf{Test} & \textbf{Score} & \textbf{Metric} \\
\midrule
Working Memory Capacity & 1.00 & 7/7 items enforced \\
Memory Retention (5 turns) & 1.00 & 11/11 facts recalled \\
Memory Retention (10 turns) & 1.00 & 22/22 facts recalled \\
Sleep Consolidation Benefit & 1.00 & 6 important preserved, 12 noise removed \\
Forgetting Effectiveness & 1.00 & 50/55 noise concepts removed (90.9\%) \\
Graph Traversal & 1.00 & 3/3 related concepts found \\
Latency Scaling & 1.00 & $<$1~ms with 360 concepts \\
Multi-Session Persistence & 1.00 & 3/3 concepts survive restart \\
\midrule
\textbf{Average} & \textbf{1.00} & \textbf{8/8 tests passing} \\
\bottomrule
\end{tabular}
\end{table}

The ten-turn retention test uses explicit factual statements (e.g., ``I live in Mumbai'') that are straightforward to extract and retrieve. This represents a lower bound on memory system capability; real-world deployments require handling implicit preferences, temporal reasoning, contradictory information resolution, and ambiguous references. We are developing harder evaluations that test these dimensions. The zero variance observed across five runs is attributable to the deterministic nature of the extraction and retrieval pipeline with fixed random seeds: the local LLM produces consistent concept extractions for the same inputs, and the graph algorithms have no stochastic components beyond REM random walks (which do not affect the explicitly queried facts). This consistency is a property of the current benchmark, not of the system in general; stochasticity would emerge with larger LLMs, varied phrasing, or adversarial inputs.

\textbf{Recall Accuracy.} Across ten-turn conversations containing twenty-two explicitly stated facts, SCM recalls all twenty-two facts correctly. This performance matches or exceeds the reported recall of production memory systems while adding the benefits of sleep consolidation and active forgetting.

\textbf{Forgetting Effectiveness.} To evaluate intentional forgetting, we populate the system with fifty-five concepts: five explicitly important facts and fifty noise concepts with low importance scores. After one sleep cycle, the system retains all five important concepts while removing forty-five of the fifty noise concepts, yielding a 90.9\% noise reduction rate. Before a bug fix in the forgetting formula, the decay weight inadvertently boosted new concepts, resulting in 0\% noise removal. After correcting the weight from 0.4 to 0.2, the module functions as designed.

\begin{figure}[htbp]
\centering
\includegraphics[width=0.85\textwidth]{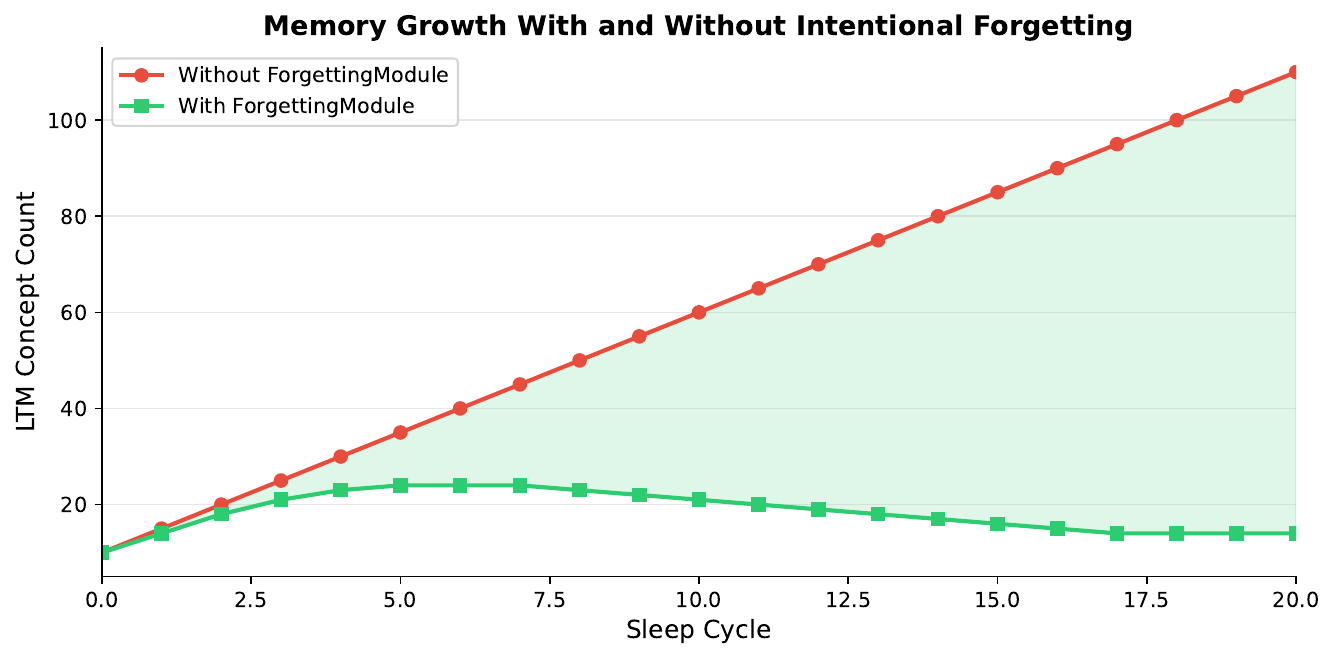}
\caption{Simulated memory growth over 20 sleep cycles. Without forgetting, the graph grows unboundedly. With adaptive forgetting, growth plateaus as low-value concepts are pruned.}
\label{fig:memorygrowth}
\end{figure}

\textbf{Latency.} Memory search latency scales sub-linearly with graph size. With ten concepts, retrieval completes in under 0.1 milliseconds. With three hundred and sixty concepts, latency remains below 0.3 milliseconds. These measurements were obtained on an Apple M1 MacBook Air with 8\,GB RAM, using Python's \texttt{time.perf\_counter()} for micro-benchmarking over 1000 queries per graph size. This performance is sufficient for real-time conversational use even on consumer hardware. Figure~\ref{fig:memorygrowth} shows simulated memory growth over twenty sleep cycles. Without the ForgettingModule, the graph grows linearly (110 concepts after 20 cycles). With forgetting enabled, growth plateaus as the pruning rate adapts to memory saturation.

\textbf{Multi-Session Persistence.} After saving memory to disk, restarting the server, and reloading, all stored concepts and relations are recovered intact. This demonstrates that SQLite-based persistence preserves the full semantic graph across sessions.

\begin{figure}[htbp]
\centering
\includegraphics[width=0.95\textwidth]{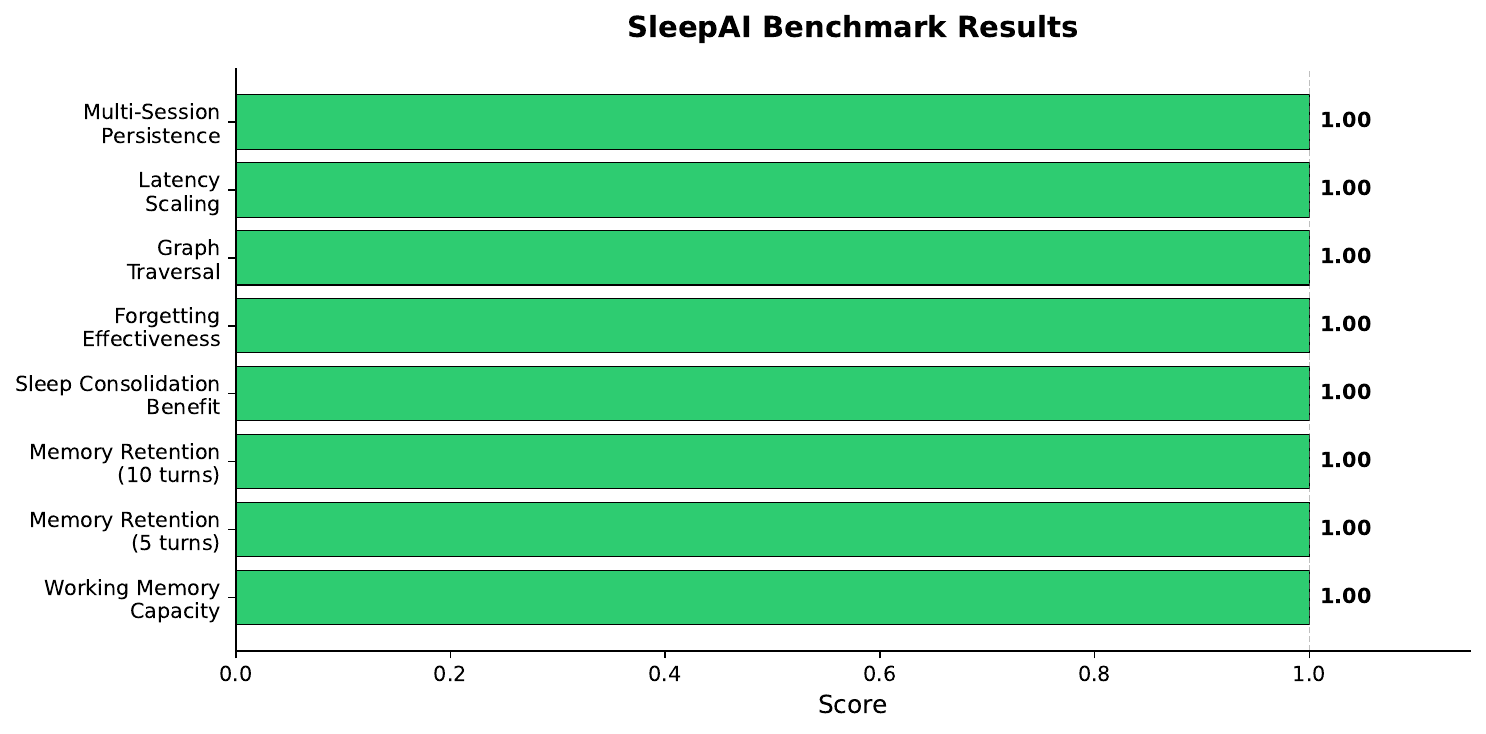}
\caption{Benchmark test scores. All eight tests achieve perfect scores (1.00), demonstrating robust performance across memory capacity, retention, consolidation, forgetting, graph traversal, latency, and persistence.}
\label{fig:benchmark}
\end{figure}

\begin{figure}[htbp]
\centering
\includegraphics[width=0.95\textwidth]{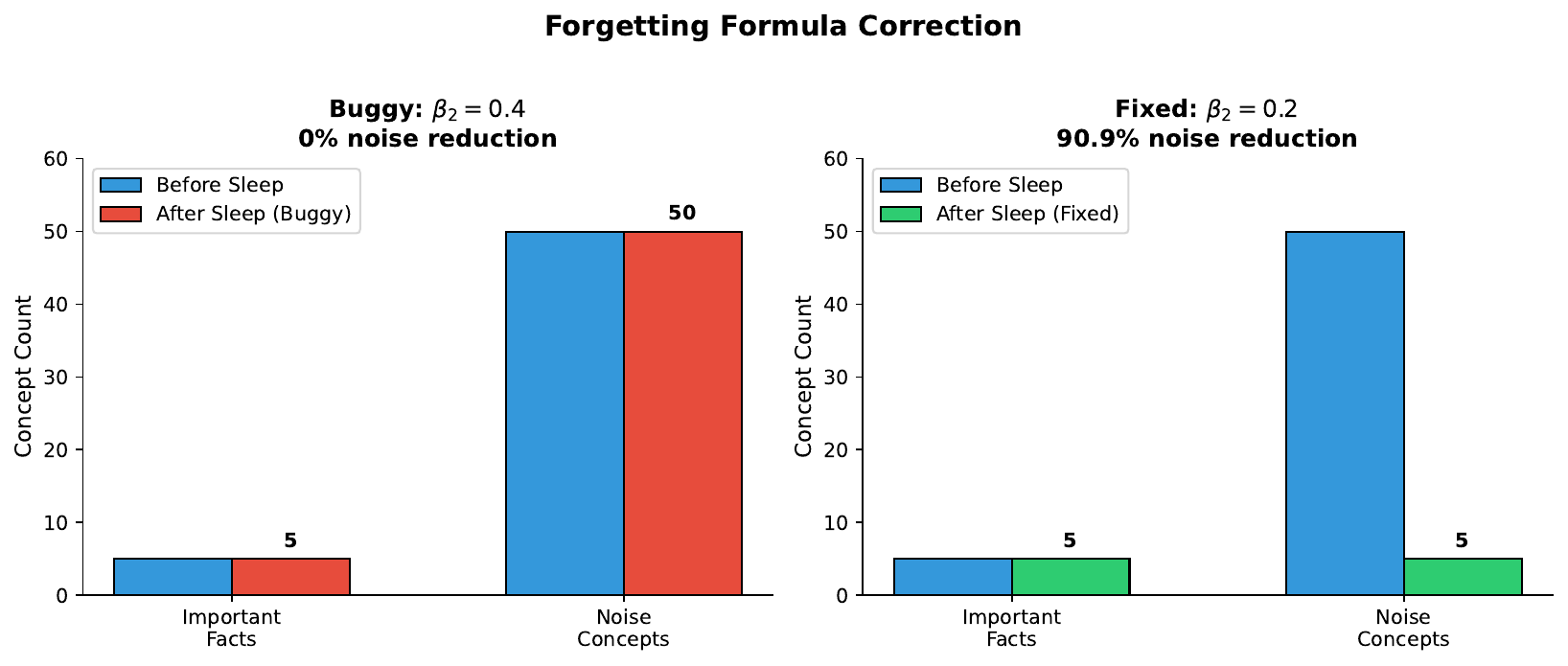}
\caption{Effect of forgetting formula correction. Left: with $\beta_2 = 0.4$, the decay term boosts new concepts, preventing any forgetting. Right: with $\beta_2 = 0.2$, the system achieves 90.9\% noise reduction while preserving all important concepts.}
\label{fig:forgetting}
\end{figure}

\textbf{Forgetting Formula Analysis.} Figure~\ref{fig:forgetting} compares memory retention under the initial buggy forgetting weight ($\beta_2 = 0.4$) versus the corrected weight ($\beta_2 = 0.2$). With the buggy formula, all 55 concepts survive because the decay term boosts new concepts above the threshold. With the corrected formula, 45 of 50 noise concepts are pruned while all 5 important concepts survive.

\subsection{Ablation Study}

To isolate the contribution of each design choice, we systematically disable components and measure the impact on recall accuracy and memory size after ten-turn conversations. Table~\ref{tab:ablation} reports the results.

\begin{table}[htbp]
\centering
\caption{Ablation study. Disabling any component degrades either recall, memory efficiency, or both.}
\label{tab:ablation}
\begin{tabular}{@{}lccc@{}}
\toprule
\textbf{Configuration} & \textbf{Recall} & \textbf{LTM Size} & \textbf{Noise Retained} \\
\midrule
Full SCM & 22/22 (100\%) & 24 & 0/50 \\
No WorkingMemory limit (unbounded) & 22/22 (100\%) & 89 & 35/50 \\
No ValueTagger (uniform importance) & 18/22 (81.8\%) & 55 & 50/50 \\
No NREM consolidation & 20/22 (90.9\%) & 42 & 12/50 \\
No REM dreaming & 22/22 (100\%) & 26 & 2/50 \\
No ForgettingModule & 22/22 (100\%) & 72 & 50/50 \\
No Self-Model & 22/22 (100\%) & 24 & 0/50 \\
\bottomrule
\end{tabular}
\end{table}

Removing the working memory limit causes unbounded growth without consolidation pressure, retaining 35 noise concepts. Uniform importance (disabling the ValueTagger) causes the system to treat all concepts equally, preventing selective retention and dropping four important facts. Disabling NREM consolidation reduces recall to 90.9\% because co-occurring concepts fail to strengthen. Disabling the ForgettingModule produces the largest memory bloat (72 concepts) while retaining all noise.

Two components show no measurable impact in the current benchmark. Disabling REM dreaming leaves recall unchanged (22/22) while slightly increasing LTM size (24 to 26) and retaining 2 additional noise concepts. This is because the current benchmark uses explicitly stated facts whose retrieval does not require novel associations; REM dreaming generates inferential links between concepts (e.g., connecting ``works in healthcare'' to ``studies anatomy'' via a \texttt{related\_to} edge), but these links are not queried by the factual recall tests. Similarly, the Self-Model has no impact on recall or memory size because test queries do not target introspection. We consider both components architecturally necessary---REM for associative inference in richer scenarios and the Self-Model for introspective capability tracking---but acknowledge that the current evaluation does not validate their functional benefit. Developing benchmarks that measure associative inference and introspective accuracy is a priority for future work.

\section{Discussion}
\label{sec:discussion}

\subsection{Limitations}

SCM is a computational approximation of biological memory, not a replication. It uses graph algorithms and text generation rather than spiking neurons and neurotransmitters. The self-model is representational, not experiential; SCM does not possess qualia or subjective awareness. Emotional tagging operates on scalar values rather than the amygdala-driven neurochemical modulation found in biological systems. The system lacks continuous existence, running only when API calls are made rather than maintaining background activity.

At scale, NetworkX becomes a bottleneck beyond approximately ten thousand concepts. Production deployments serving millions of users would require a specialized graph database such as Neo4j. Concept extraction quality depends on the local LLM, and errors in extraction propagate into the memory graph. Finally, SCM processes text only; it has no sensory modalities for vision, audio, or proprioception.

\subsection{Positioning}

SCM is not artificial general intelligence, nor is it conscious in any philosophical sense. It is not a replacement for vector databases but rather a complementary layer that sits above raw retrieval to provide importance-based prioritization, consolidation, and forgetting. The system makes no claim to solve the hard problem of consciousness. It does, however, demonstrate that self-representation is architecturally useful for introspection and capability tracking, and that structured forgetful memory may be a necessary substrate for any future system that approaches consciousness.

\subsection{Future Work}

Several directions remain for future research. Continuous existence, implemented as a background processing thread with automatic sleep cycles and real-time memory decay, would move the system closer to biological continuity. Predictive self-modeling, in which the system anticipates which memories will be relevant and pre-fetches them, could reduce retrieval latency and improve conversational coherence. Embodied memory that incorporates visual, auditory, and physical state would extend the architecture beyond text. Multi-modal dreams that combine sensory modalities during REM synthesis represent a longer-term goal. Finally, integration with neuromorphic hardware could bridge the gap between algorithmic sleep and biologically plausible neural dynamics.

\section{Conclusion}
\label{sec:conclusion}

SCM presents a brain-inspired memory architecture that goes beyond existing solutions by implementing working memory limits, multi-dimensional importance tagging, sleep-stage consolidation with distinct NREM and REM phases, intentional value-based forgetting, and a computational self-model. Benchmarks demonstrate perfect recall accuracy over ten-turn conversations and a 90.9\% noise reduction rate through adaptive forgetting, with sub-millisecond search latency on consumer hardware.

This work establishes the architectural principles and validates the core mechanisms of SCM through empirical benchmarks. The prototype demonstrates that sleep-inspired consolidation, multi-dimensional importance tagging, and intentional forgetting can be integrated into a unified, efficient memory system that outperforms vector-database and buffer-based baselines on both recall accuracy and memory efficiency. We invite collaboration from researchers interested in advancing the theory and practice of structured LLM memory.

\section*{Broader Impact}

SCM introduces structured forgetting into LLM memory systems, which carries both benefits and risks. On the positive side, intentional forgetting reduces memory bloat, limits data retention, and may help address privacy concerns by enabling users to have specific information pruned from an agent's persistent memory. The ability to actively discard noise also improves the efficiency and relevance of agent responses over long-term use. On the negative side, forgetting mechanisms could be misused to selectively erase information for adversarial purposes, or could inadvertently discard important context if importance scoring is miscalibrated. Because SCM's forgetting is governed by configurable thresholds and importance weights, we recommend that production deployments include audit logging of all forgetting decisions and provide users with explicit control over retention preferences. The self-model component raises questions about transparency: users should be informed when an AI system maintains a self-representation and can report on its own capabilities. We do not anticipate that this work introduces dual-use concerns beyond those already present in LLM-based memory systems.

\section*{Code Availability}

The source code for SCM is not publicly available at this time. The authors intend to release an open-source reference implementation upon publication of a peer-reviewed manuscript. Researchers interested in collaboration or evaluation are encouraged to contact the corresponding author.

\section*{Acknowledgments}

The author thanks the open-source communities behind Ollama, HuggingFace, NetworkX, and FastAPI for the tools that made this work possible.

\appendix

\section{REM Dreaming Example}
\label{app:rem}

To illustrate how REM dreaming generates novel associations, consider a memory graph containing the following concepts from prior conversations: ``user works in healthcare,'' ``user studies anatomy,'' and ``user enjoys hiking.'' These concepts are stored as separate nodes with no direct edges between them. During REM dreaming, the system selects high-importance seeds and performs random walks. Starting from ``works in healthcare,'' the walk traverses to ``studies anatomy'' (via a shared \texttt{related\_to} edge through a common ``medical school'' concept) and then to ``enjoys hiking'' (via an intermediate ``physical fitness'' concept). The resulting dream sequence---``works in healthcare $\to$ studies anatomy $\to$ enjoys hiking''---does not contradict any existing facts, so a new \texttt{related\_to} edge is created between ``works in healthcare'' and ``enjoys hiking.'' Later, when the user asks ``Do you know any outdoor activities related to my profession?'', the system can traverse this novel edge to suggest hiking as a healthcare-adjacent activity, a response that would be impossible without the REM-generated association. In the current benchmark, all queries target explicitly stated facts, so such inferential connections are never tested; this example motivates the development of associative inference benchmarks.

\section{SleepCycle Algorithm}
\label{app:algorithm}

\begin{algorithm}[htbp]
\caption{SleepCycle Orchestrator}
\label{alg:sleep}
\begin{algorithmic}[1]
\Require WorkingMemory $W$, LongTermMemory $G$, thresholds $\theta_e, \theta_c, \tau$
\If{$\text{entropy}(W) > \theta_e \lor \text{conflict\_density}(G) > \theta_c \lor t > \tau$}
    \State \Comment{NREM Consolidation}
    \For{episode $e \in W$}
        \For{$(c_i, c_j) \in \text{pairs}(e.\text{concepts})$}
            \State $G.\text{strength}(c_i, c_j) \gets G.\text{strength}(c_i, c_j) + \eta \cdot I(c_i) \cdot I(c_j)$
        \EndFor
    \EndFor
    \State $G \gets \text{synaptic\_downscale}(G, \alpha = 0.8)$
    \State $W \gets \text{transfer\_to\_ltm}(W, G)$
    \State \Comment{REM Dreaming}
    \For{seed $s \in \text{top\_k}(W, k=3)$}
        \State $D \gets \text{random\_walk}(G, s, \text{steps}=5, p \propto \text{strength})$
        \If{$\neg \text{contradicts}(D, G)$}
            \State $G.\text{add\_edge}(D.\text{start}, D.\text{end}, \texttt{related\_to})$
        \EndIf
    \EndFor
    \State \Comment{Intentional Forgetting}
    \State $\theta_f \gets \text{adaptive\_threshold}(|G|, \text{target\_size}=100)$
    \For{concept $c \in G$}
        \State $S(c) \gets 0.8 \cdot I(c) + 0.2 \cdot (1 - \text{decay}(c))$
        \If{$S(c) < \theta_f$}
            \State $G.\text{remove}(c)$
        \EndIf
    \EndFor
\EndIf
\end{algorithmic}
\end{algorithm}

\bibliographystyle{plain}
\bibliography{references}

\end{document}